 \documentclass[pmlr,twocolumn,10pt]{jmlr} 

\usepackage{lscape}
\usepackage{booktabs}
\usepackage[load-configurations=version-1]{siunitx} 

\newcommand{\equal}[1]{{\hypersetup{linkcolor=black}\thanks{#1}}}

\theorembodyfont{\upshape}
\theoremheaderfont{\scshape}
\theorempostheader{:}
\theoremsep{\newline}

\jmlrvolume{LEAVE UNSET}
\jmlryear{2021}
\jmlrsubmitted{LEAVE UNSET}
\jmlrpublished{LEAVE UNSET}
\jmlrworkshop{Machine Learning for Health (ML4H) 2021} 

\title[Invariant Risk Minimisation for Cross-Organism Inference]{Invariant Risk Minimisation for Cross-Organism Inference: \titlebreak Substituting Mouse Data for Human Data in Human Risk Factor Discovery}

 \author{%
 \Name{Odhran ODonoghue} \equal{These authors contributed equally} \Email{odhran.odonoghue@eng.ox.ac.uk} \addr Oxford University\\
 \Name{Paul Duckworth } \footnotemark[1] \Email{pduckworth@robots.ox.ac.uk} \addr Oxford University\\
   \Name{Giuseppe Ughi} \footnotemark[1] \Email{ughi@maths.ox.ac.uk} \addr Oxford University\\
   \Name{Linus Scheibenreif} \footnotemark[1] \Email{linus.scheibenreif@unisg.ch} \addr  University of St Galen\\
   \Name{Kia Khezeli} \Email{kk839@cornell.edu} \addr Mayo Clinic\\
   \Name{Adrienne Hoarfrost } \Email{adrienne.l.hoarfrost@gmail.com} \addr NASA \\
   \Name{Samuel Budd} \Email{budd.samuel@gmail.com} \addr King's College London\\
   \Name{Patrick Foley} \Email{patrick.foley@intel.com} \addr Intel Corporation\\
   \Name{Nicholas Chia } \Email{Chia.Nicholas@mayo.edu} \addr Mayo Clinic\\
   \Name{John Kalantari} \Email{Kalantari.John@mayo.edu} \addr Mayo Clinic\\
   \Name{Graham Mackintosh} \Email{graham.mackintosh@nasa.gov} \addr NASA\\
   \Name{Frank Soboczenski} \Email{frank.soboczenski@kcl.ac.uk} \addr King's College London\\
   \Name{Lauren Sanders} \Email{lauren.m.sanders@nasa.gov} \addr NASA\\
  }

\begin{document}

\maketitle

\begin{abstract}
Human medical data can be challenging to obtain due to data privacy concerns, difficulties conducting certain types of experiments, or prohibitive associated costs. 
In many settings, data from animal models or \textit{in-vitro} cell lines are available to help augment our understanding of human data. 
However, this data is known for having low etiological validity in comparison to human data. 
In this work, we augment small human medical datasets with \textit{in-vitro} data and animal models. 
We use Invariant Risk Minimisation (IRM) to elucidate invariant features by considering cross-organism data as belonging to different data-generating environments. 
Our models identify genes of relevance to human cancer development. 
We observe a degree of consistency between varying the amounts of human and mouse data used, however, further work is required to obtain conclusive insights. 
As a secondary contribution, we enhance existing open source datasets and provide two uniformly processed, cross-organism, homologue gene-matched datasets to the community. 
\end{abstract}


\section{Introduction}
This work is motivated by the biological challenge of spaceflight and the effects of cosmic radiation on human health. Cosmic radiation is able to penetrate thick layers of shielding and body tissue and its carcinogenic nature is a major cause for concern for long-distance space travel~\citep{chancellor2014space}. Understanding the exact mechanisms of radiation damage could give us insight into how to mitigate its effects on cells and tissues~\citep{durante2011physical}. However, studying these mechanisms is difficult because human space data is incredibly expensive and difficult to generate~\citep{haukkala2017financing}. As a result, studies on human data typically do not have the statistical power required to identify risk factors - particularly in high-dimensional genomics and transcriptomics datasets \citep{wheelock2013trials}. 

\textit{In-vitro} experiments on human cells are an alternative data source, but these lack the etiological validity of a complete organism \citep{gillette1984problems} - especially given how radiation interacts with the different tissues it is penetrating \citep{parihar2016cosmic}.  
A second alternative is animal models, which can be exposed to radiation that mimics a cosmic profile. While this data is much easier to generate, there is a poor record of experiments conducted in animal models translating into useful human findings due to the obvious dissimilarities between organisms \citep{bracken2009animal}. 
This abundance of animal model data but a dearth of human data is by no means unique to cosmic radiation experiments: it is an issue encountered routinely for conditions that are hard to experimentally study in humans without putting them at risk\citep{martic2012can, betarbet2002animal}. 

A causal framework has been proposed in the literature based on identifying invariance across different data-generating environments, leading to advances in domain generalization~\citep{cartwright2003two,arjovsky2019invariant,buhlmann2020}\footnote{The choice of IRM in this work has an exploratory purpose; these results suggest that other causal methods would also work. }. 
Environments are defined as subsets of available data that do not share the same underlying populations, 
but for which the causal relationships to a target variable are assumed invariant, 
e.g. different hospitals might capture data about a disease but with varying populations of underlying health characteristics. 

Our contributions include demonstrating the effectiveness of this framework for identifying invariant relationships present in cross-organism datasets. 
Further, we provide two open source cross-organism datasets to the community for further research (with matched gene-homologues): one based on acute gamma radiation experiments of humans and mouse data, and the second on chronic heavy ion radiation experiments. 
%
Our results provide valuable insights into a set of human-relevant health variables in a setting where there is a deficit of human data but an abundance of model organism data. 

\section{Invariant Risk Minimisation} 
Invariant Risk Minimization (IRM) is designed to elucidate invariant relationships from empirical data. In particular, IRM is a learning method that seeks to identify a data representation equally optimal across the different environments. More precisely, IRM is expressed as a constrained optimization problem of the following form: 

\begin{align}
    &\min_{\substack{\Phi: \mathcal{X}\rightarrow \mathcal{H}\\w: \mathcal{H}\rightarrow \mathcal{Y}}} \qquad \sum_{e\in \mathcal{E}_{\mathsf{tr}}} R^e\left(w \circ \Phi\right) \label{eq:IRM}\\
    &\mathrm{subject~to}\quad w\in \underset{\Tilde{w}:\mathcal{H}\rightarrow \mathcal{Y}}{\mathrm{argmin}}\  R^e\left(\Tilde{w} \circ \Phi\right),\ \forall e\in \mathcal{E}_{\mathsf{tr}}, \nonumber
\end{align}

where $\mathcal{E}_{\mathsf{tr}}$ is the set of training environments, $R^e$ denotes the risk under environment $e$, $\Phi$ is a learnable data representation and $w$ is a fixed classifier. 
Notice that removing the constraint in Equation~\eqref{eq:IRM} recovers the classical empirical risk minimization problem. 
Incorporating this additional environment-dependant constraint results in a bi-leveled optimization problem, which is computationally challenging. 
\cite{arjovsky2019invariant} proposes the following variant of IRM that is more practical: 

\begin{align}
    \min_{\substack{\Phi: \mathcal{X}\rightarrow \mathcal{H}}} \qquad \sum_{e\in \mathcal{E}_{\mathsf{tr}}} &R^e\left(w_0 \circ \Phi\right) \label{eq:IRMv1}\\&+ \lambda \left\| \nabla_{w|w=w_0} R^e\left(w\circ \Phi\right) \right\|^2, \nonumber 
\end{align}

where $w_0$ is a user specified vector. 
Furthermore, the choice of IRM in the original paper \citep{arjovsky2019invariant} assumes that relationships between predictors and outcomes are linear. However, the model $\Phi$ can be chosen as a general multi-layer Perceptron to model non-linear relations, as in~\citep{lu2021nonlinear}.

In order to confirm that our implementation of IRM performs as intended on real biological datasets, we used a synthetic dataset where the causal structure is known in advance for calibration. The synthetic dataset was generated using structural causal models (SCMs) ~\cite{pearl1995causal} as proposed in recent literature ~\citep{aubin2021linear}. As shown initially by \citep{arjovsky2019invariant} on these data, classifiers trained with IRM lead to significantly higher test accuracy in the test cases than with traditional methods based on empirical risk minimisation, especially on classification tasks which is a similar configuration to our biological setting (we found up-to a 5X improvement in test set accuracy). Furthermore, the risks that IRM is susceptible to \cite{rosenfeld2020risks} did not manifest themselves; thus suggesting that IRM should not encounter majour issues in our setting.  

\section{Cross-Organism Datasets}\label{sec:data}

\subsection{Data}

Multiple radiation exposure datasets were used from the publicly accessible NASA GeneLab data repository of human, animal and cell line experiments (available at~\citep{nasa-genelab}). We identified datasets from individual experiments with similar radiation exposure levels and duration.
Through a set of pre-processing steps detailed below, we combined individual mouse and human experiments to create larger, uniformly processed and gene-mapped datasets. 
The merged cross-organism datasets, including all pre-processing steps, are available open source to the research community (link removed for anonymity). 


\paragraph{Dataset 1 Acute Gamma Radiation:}
The first datasets selected were experiments of messenger RNA (mRNA) sequencing from blood cell populations where either human blood samples or mouse models were exposed to acute gamma radiation. This consisted of two human experiments (GLDS-157, GLDS-152; total humans=125)~\citep{GLDS-157, GLDS-152} and three mouse experiments (GLDS-156, GSE124612, GSE62623; total mice=296)~\citep{paul2019transcriptomic, GLDS-156, paul2015radiation}. 

\paragraph{Dataset 2 Chronic Heavy Ion:}
For our second dataset we identified experiments of mRNA sequencing from lung tissues in which either human lung tissue cell lines or mouse models were exposed to chronic heavy ion radiation (which more closely mimics a cosmic radiation profile than gamma rays\citep{buecker1975biological}). These experiments include one human bronchial cell experiment  (GLDS-73; total cell culture samples=95)~\citep{GLDS-73} and one mouse lung experiment (GLDS-148; total mice=41)~\citep{GLDS-148}. 

\subsection{Pre-Processing}
To make the human and mouse datasets suitable for our IRM pipeline, we performed the following pre-processing steps. We identify a pool of gene homologues (i.e. genes with similar structure and function) between the organisms, using the \textit{Ensembl} database~\citep{howe2021ensembl}, maintained by the European Molecular Biology Laboratory. 
Using this database we convert suitable mouse genes into their human-homologue equivalents~\footnote{Note: occasional poor homologue matches are not a cause for concern, as if there is a significant variance between human gene behaviour and mouse homologue behaviour, this feature will be downweighted by IRM as the relationship will not be invariant between environments.}. 
This pre-processing results in the loss of some genes where homologues cannot be found.

Secondly, we performed a dimensionality reduction to remove the genes with lowest variance across the combined dataset. 
In the following experiments, the top $1000$ genes with the highest variance were kept for analysis. 
Data from each experiment were normalised to a $Z$-distribution. Following this, data were sorted into environments, each corresponding to its individual human or mouse original experiment. 
The environments are used in the IRM algorithm to elucidate the invariant features and identify potential causal drivers of the target variable. 

\section{Experiments and Results:}
\subsection{Experimental Setup}

We set up our experiment as a binary classification task and trained an IRM model to classify irradiated samples from non-irradiated controls. We consider two types of experiments (augmentation and substitution) for each combined biological dataset presented in Section~\ref{sec:data}. In \textbf{Augmentation Experiments}  we investigate varying the amount of mouse samples available in the cross-organism dataset. Mouse samples were added to the original human samples in small increments until all mouse samples were combined with the human samples. In \textbf{Substitution Experiments} we substituted human data samples with mouse samples on an incremental basis. That is, we begin with all human data and two mouse samples per environment, and incrementally replace the human samples with mouse samples until only two human samples per environment remain.

For each experiment, the output is a ranked list of features (genes), where the order is defined by 
coefficients extrapolated from the model $\Phi$ in Equation~\ref{eq:IRMv1}; the more a feature is consistently predictive of the target variable across multiple environments, the higher the corresponding coefficient will be.
These \textit{invariant} features are then considered candidates for causal features.  
In each experiment, we are interested to evaluate the change in the reported highest ranked features as the mouse-human ratio changes. Thus, we compare the gene rankings that IRM returns for each combination of mouse-human data in a pairwise fashion. Specifically, we consider how many of the top-10, or top-50 ranked genes overlap between each mixed-organism experiment. We also consider two similarity metrics common in the literature: the Ranked-Biased overlap~\citep{webber2010similarity} of the ranked features and their Kendal-tau similarity metric~\citep{Puka2011}. 


\begin{figure*}[htbp]
\centering
\begin{minipage}{.48\textwidth}
  \centering
  \includegraphics[width=0.9\linewidth, trim=0.23cm 0.23cm 0.2cm 0.2cm,clip]{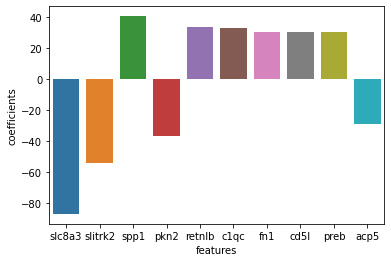}
  \vspace{-0.3cm}
  \caption{Top-10 IRM features according to feature coefficients with all human and mice gamma-radiation data.}
  \label{fig:gamma_ray_top}
\end{minipage}%
\hfill
\begin{minipage}{.48\textwidth}
  \centering
  \includegraphics[width=0.9\linewidth, trim=0.23cm 0.23cm 0.2cm 0.2cm,clip]{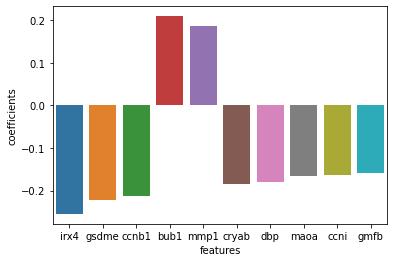}
  \vspace{-0.3cm}
  \caption{Top-10 IRM features according to feature coefficients with all human and mice heavy-ion radiation data.}
  \label{fig:heavy_ion_top}
\end{minipage}
\vspace{-0.5cm}
\end{figure*}

\subsection{Biomedical Findings}

\textbf{Dataset 1 Acute Gamma Radiation: } Appendix figures show 4 similarity metrics for the augmentation experiment on different mixed-organism experiments. In particular, the Ranked-Bias overlap similarity is consistently above 0.5 correlation for all comparisons, demonstrating strong overlap in the IRM reported highest ranked features. As expected, there is more overlap in the top-50 features than the top-10, with at least $10\%$ overlapping in most comparisons. We also investigated the biological relevance of the top-10 highest ranked genes (Figure~\ref{fig:gamma_ray_top}). In experiments containing all mouse and human data IRM consistently identifies several genes known to be relevant to cancer pathophysiology. For example, PKN2, RETNLB, and SPP1~\citep{lin2017protein, jin2021investigating, xu2017spp1}. Of particular interest is the gene with the highest coefficient, \textit{SLC8A3}. This is a relatively under-studied protein that is involved in Sodium-Calcium exchange~\citep{gabellini2002human}. 
This result suggests there is value in investigating the function of this compound to understand its role in pathophysiological pathways outside of its currently implicated role in arthritis~\citep{julia2016genome}. 

\textbf{Dataset 2 Chronic Heavy Ion:} Experiments containing all mouse and human data identified \textit{IRX4}, \textit{GSDME}, \textit{CCNB1}, \textit{BUB1}, and \textit{MPP1} as the 5 genes with highest coefficients (Figure~\ref{fig:heavy_ion_top}). 
Each one of these genes are directly implicated in the literature on the pathophysiology of Lung cancer~\citep{wang2020comprehensive, zhang2019chemotherapeutic,  wang2019degradation, haruki2001molecular, sauter2008matrix}. Our findings demonstrate that IRM is able to provide powerful insights into biological mechanisms relevant to human pathophysiology, and identify potentially causal genes from observational datasets. 



Across the experiments, we found that overlap is observed in the top-10, and top-50 features out of 1000 features selected by the model across different divisions of mouse and human data. 
However, Ranked Bias Overlap Scores and Kendal-tau scores indicate there is variance between the results of models with different number of mice and human data points, as shown in Figure~\ref{fig:heatmap}. It is difficult to directly interpret if this affects the validity of our findings as the underlying causal structure of the interactions studied in this work are unknown. 
Nevertheless, further exploration is needed. 
One line of future work could be to use alternative domain-generalisation methods 
such as Invariant Causal Prediction (ICP)~\citep{heinze-deml2017nvariant}, however ICP does not scale well to large numbers of features.  

\vspace{-0.4cm}
\section{Conclusion and future work}

We demonstrate a novel paradigm for generating human-relevant biomedical insights from observational datasets with limited size by leveraging and augmenting with animal model data. 
Experiments are presented in radiation exposure and carcinogenics,
with a key novel contribution being the identification of \textit{SLC8A3} as a potentially causal feature.
This paradigm also has utility across many biomedical settings, e.g. when combining data from \textit{in-vivo} experiments in living whole organisms with \textit{in-vitro} experiments of cell and tissue cultures. 

As a second contribution we provide  two uniformly processed, cross-organism, homologue gene-matched datasets to the community . 
While our work is principally a proof-of-concept, we believe our findings and insights indicate that environment-based invariant machine learning approaches will have significant utility for medical data challenges indicating a fruitful research avenue for future work.

\acks{The authors gratefully acknowledge funding support from NASA Biological and Physical Sciences, NASA Human Research Program, and Intel. The authors would also like to thank the Frontier Development Lab and the SETI Institute for their continuous support during this work. The authors would also like to thank the NASA Ames Research Center, Intel and the Mayo Clinic for their continuous support throughout this project. The authors are also grateful for the amazing computational resources provided by Google Cloud.}

\bibliography{jmlr-sample}




\appendix

\section{Data Used}


The data used in this study can be found at: \url{https://drive.google.com/file/d/1fCclY7Cgyor8_Bbz6m2OZutzNIVPqD7u/view?usp=sharing}

\section{Figures}

\begin{figure*}[h!]
    \vspace{0.1cm}
    \centering
    \includegraphics[width=1\textwidth, ]{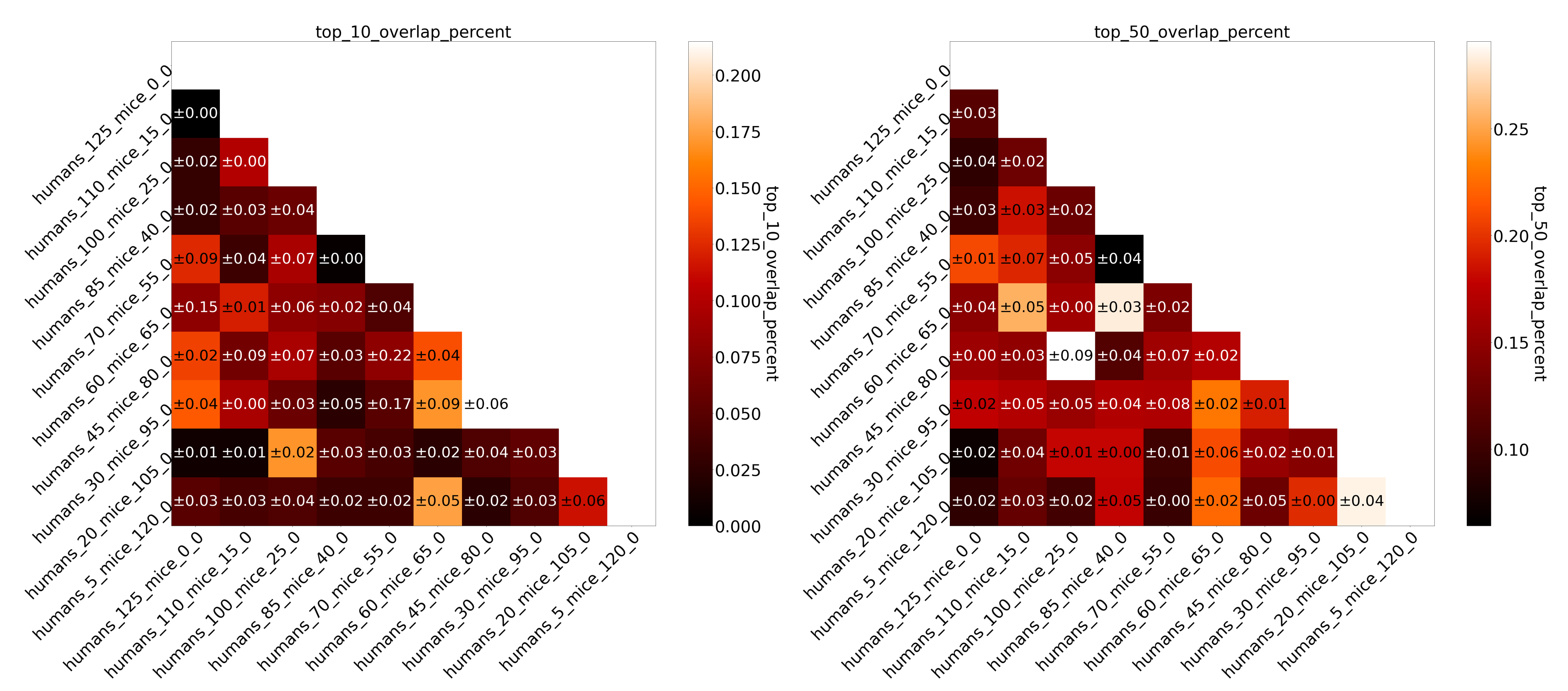}
    \vspace{-0.3cm}
    \caption{Similarity of the top-10 and top-50 overlap coefficients in the augmentation experiment using the gamma-ray exposure combine dataset. The number of samples is kept fixed to 125 and the ratio of human-to-mouse data varies. For each metric we plot the heatmap of the different possible combinations reporting inside the $95\%$ confidence interval.}
    \vspace{-0.1cm}
    \label{fig:heatmap}
\end{figure*}

\begin{figure*}[h!]
    \vspace{0.1cm}
    \centering
    \includegraphics[width=1\textwidth, ]{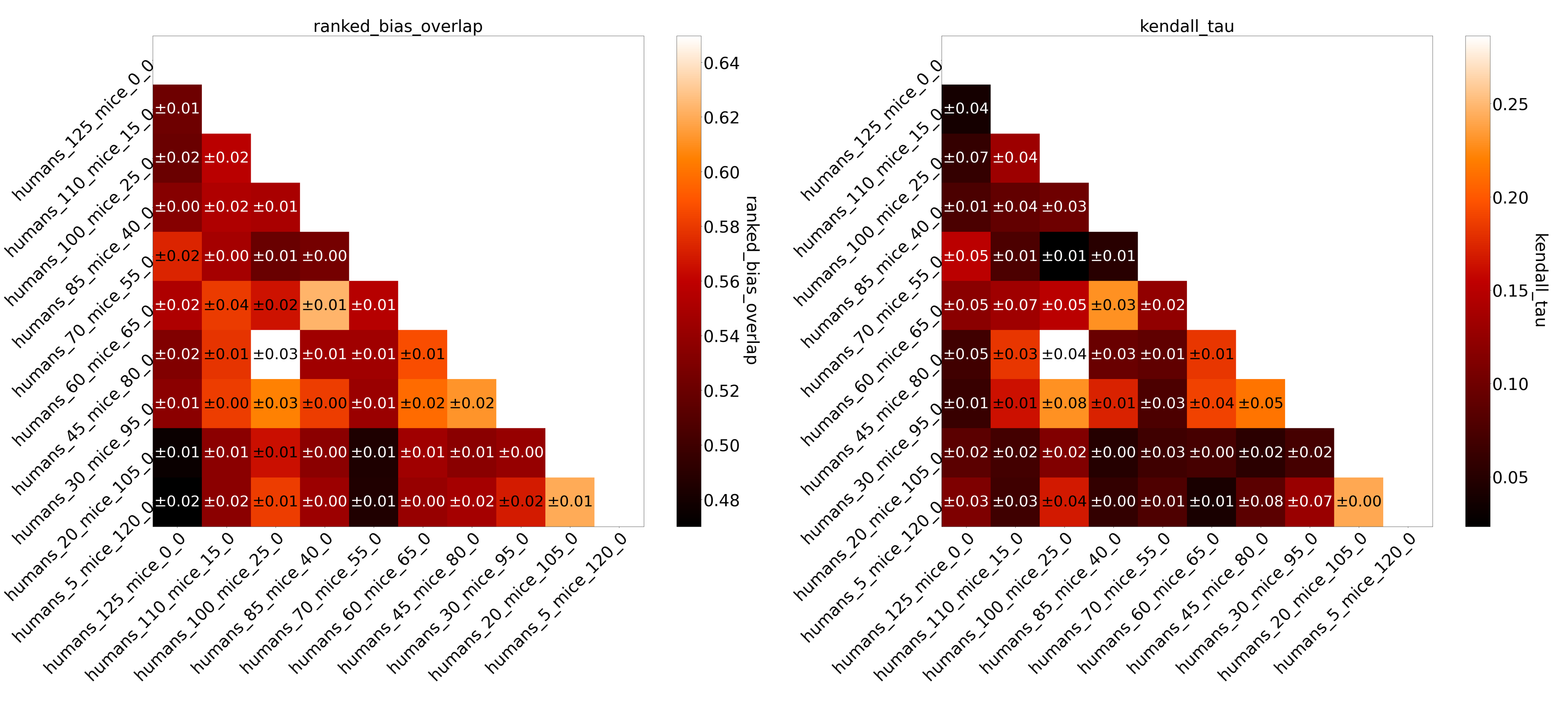}
    \vspace{-0.3cm}
    \caption{Similarity of the Ranked Bias Overlap and Kendall-Tau coefficients in the augmentation experiment using the gamma-ray exposure combine dataset. The number of samples is kept fixed to 125 and the ratio of human-to-mouse data varies. For each metric we plot the heatmap of the different possible combinations reporting inside the $95\%$ confidence interval.}
    \vspace{-0.1cm}
    \label{fig:heatmap}
\end{figure*}

\begin{figure*}[h!]
    \vspace{0.1cm}
    \centering
    \includegraphics[width=1\textwidth, ]{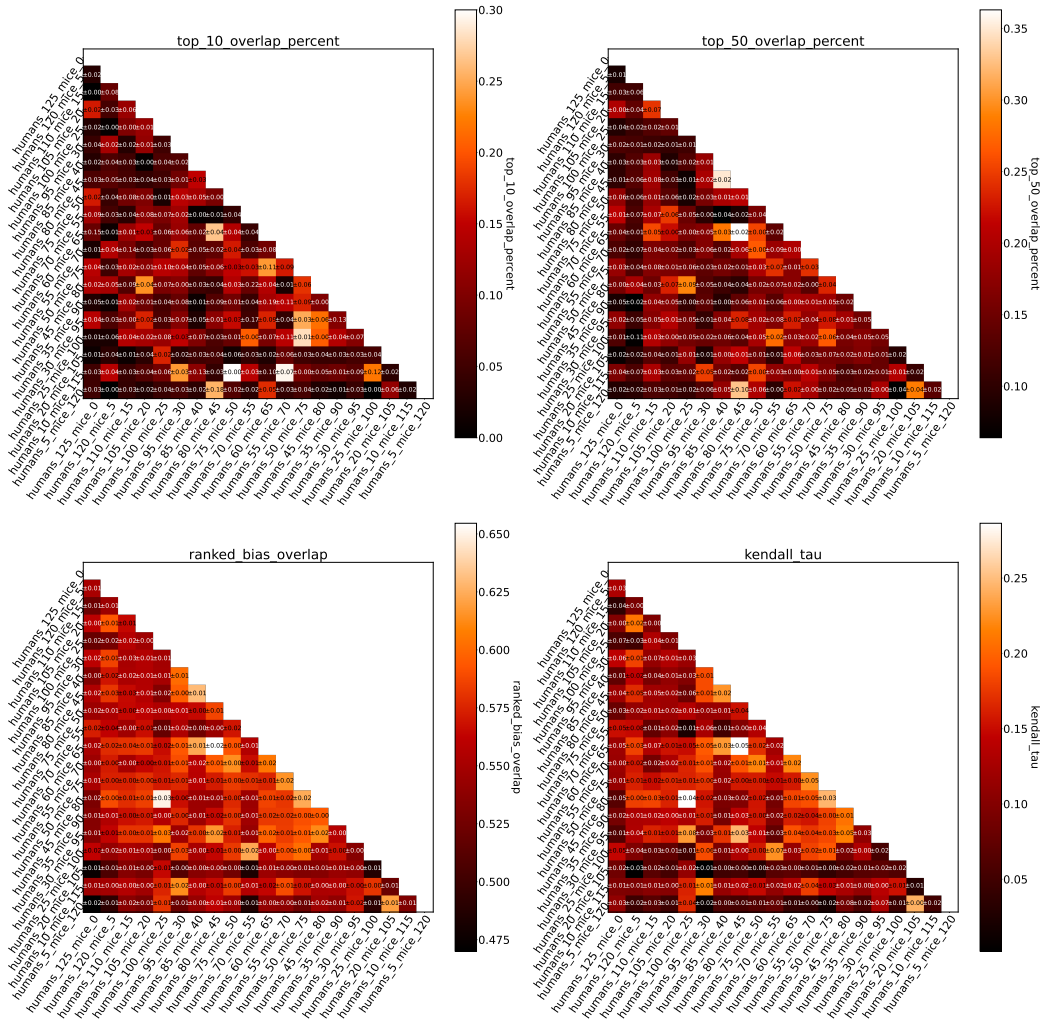}
    \vspace{-0.3cm}
    \caption{Granular analysis of similarity of coefficients in substitution experiments for gamma ray exposure data}
    \vspace{-0.1cm}
    \label{fig:heatmap}
\end{figure*}

\begin{figure*}[h!]
    \vspace{0.1cm}
    \centering
    \includegraphics[width=1\textwidth, ]{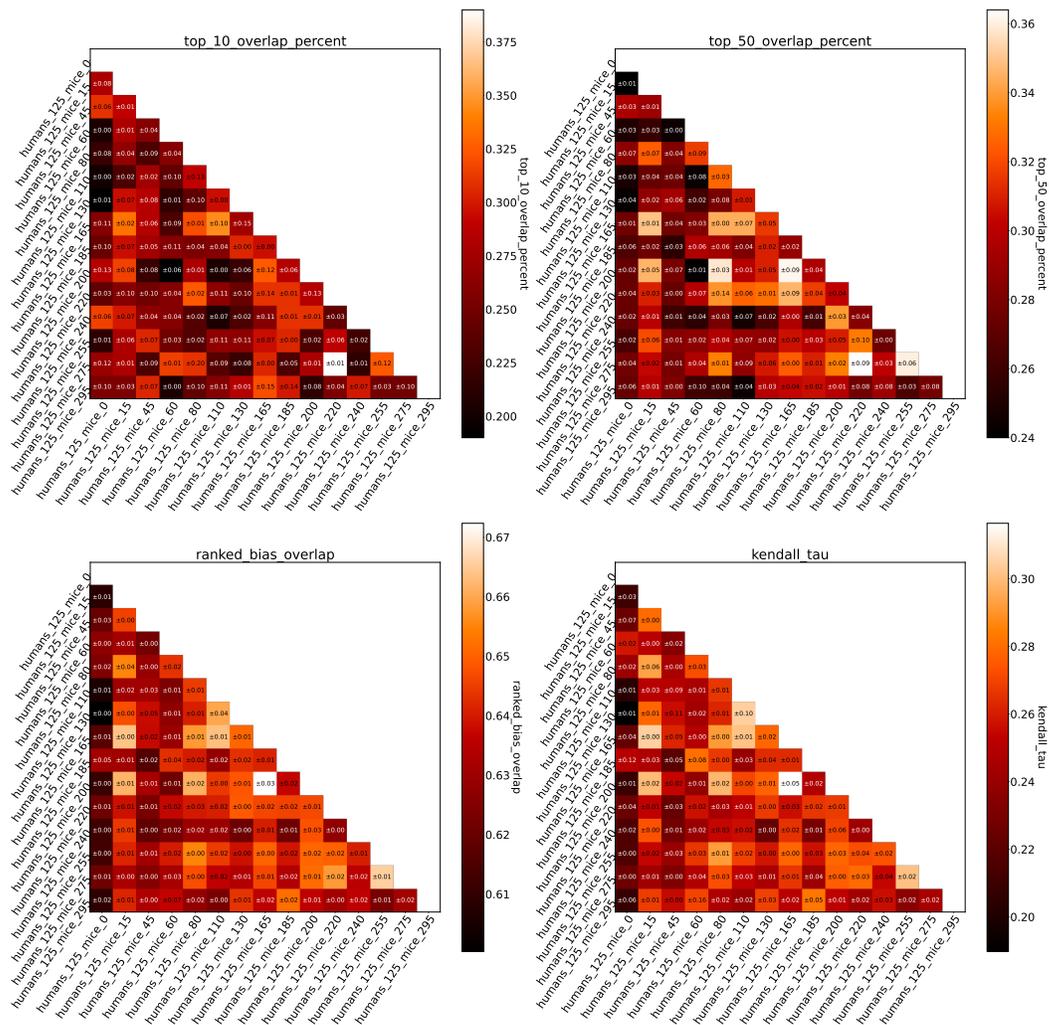}
    \vspace{-0.3cm}
    \caption{Granular analysis of similarity of coefficients in augmentation experiments for gamma ray exposure data}
    \vspace{-0.1cm}
    \label{fig:heatmap}
\end{figure*}

\end{document}